\documentclass[runningheads]{llncs}
\usepackage{graphicx}
\usepackage{subfig}

\usepackage{tikz}
\usepackage{comment}
\usepackage{amsmath,amssymb} 
\usepackage{color}

\usepackage{algorithm}
\usepackage{algpseudocode}
\usepackage[pagebackref,breaklinks,colorlinks]{hyperref}
\usepackage{multirow}
\usepackage{capt-of}    
\usepackage{tabularx}  
\usepackage{xcolor}
\usepackage{enumitem}
\usepackage{makecell}
\usepackage{bm}

\usepackage{graphicx}
\usepackage{booktabs}

\usepackage{pifont}
\newcommand{\cmark}{\ding{51}}
\newcommand{\xmark}{\ding{55}}

\usepackage{cleveref}
\crefformat{section}{\S#2#1#3}
\crefformat{subsection}{\S#2#1#3}
\crefformat{subsubsection}{\S#2#1#3}

\usepackage{amsmath}

\newcommand{\argmin}{\arg\!\min}
\newcommand{\argmax}{\arg\!\max}

\newcommand{\ie}{\textit{i}.\textit{e}. }
\newcommand{\eg}{\textit{e}.\textit{g}. }

\newcolumntype{C}[1]{>{\centering\let\newline\\\arraybackslash\hspace{0pt}}m{#1}}

\usepackage[accsupp]{axessibility}  

\usepackage{algpseudocode}

\usepackage{multirow}

\usepackage{cite}

\usepackage{mathtools}

\usepackage[misc,geometry]{ifsym}

\begin{document}
\pagestyle{headings}
\mainmatter
\def\ECCVSubNumber{4688}  

\title{FairGRAPE: Fairness-aware GRAdient Pruning mEthod for Face Attribute Classification} 

\titlerunning{FairGRAPE Pruning}
%
\author{Xiaofeng Lin
\and
Seungbae Kim
\and
Jungseock Joo\thanks{To appear in ECCV 2022.}
}
\authorrunning{X. Lin et al.}
%
\institute{University of California, Los Angeles, Los Angeles, CA 90024, USA
\email{bernardo1998@g.ucla.edu},
\email{sbkim@cs.ucla.edu},
\email{jjoo@comm.ucla.edu}
}
\maketitle

\begin{abstract}
Existing pruning techniques preserve deep neural networks' overall ability to make correct predictions but could also 
amplify hidden biases during the compression process. We propose a novel pruning method, Fairness-aware GRAdient Pruning mEthod (FairGRAPE), that minimizes the disproportionate impacts of pruning on different sub-groups. Our method calculates the per-group importance of each model weight and selects a subset of weights that maintain the relative between-group total importance in pruning. The proposed method then prunes network edges with small importance values and repeats the procedure by updating importance values. We demonstrate the effectiveness of our method on four different datasets, FairFace, UTKFace, CelebA, and ImageNet, for the tasks of face attribute classification where our method reduces the disparity in performance degradation by up to 90\% compared to the state-of-the-art pruning algorithms. Our method is substantially more effective in a setting with a high pruning rate (99\%). The code and dataset used in the experiments are available at \url{https://github.com/Bernardo1998/FairGRAPE}
\end{abstract}

\section{Introduction}


Deep neural networks (DNNs) are widely used in applications running on mobile or wearable devices where computational resources are limited~\cite{zhang2019deep}.  
A common strategy to improve the inference efficiency of deep models in such environments is model compression by pruning and 
removing insignificant nodes or connections between nodes, resulting in sparser networks than the original ones~\cite{blalock2020statepruning,frankle2018lottery,frankle2019stabilizing,han2015deep,lee2018snip,wang2020grasp}. These methods have been known to reduce the computational cost significantly with almost negligible loss in prediction accuracy~\cite{cheng2017surveydeepcompression}.

Despite the prevalence of model compression, recent studies have also reported that compressed models may suffer from hidden biases, \ie accuracy disparity, more severely than the original models~\cite{hooker2019compressed,blakeney2021simon,hooker2020bias}. The pruned models may be accurate overall or on some sub-groups (\eg White males), while resulting more severe performance decrease from the original model on specific sub-groups. This bias is particularly problematic for model pruning methods, which attempt to identify and remove insignificant parameters. The parameter-wise significance considered in such methods is estimated from in-the-wild datasets, which are typically unbalanced and biased~\cite{karkkainen2019fairface,wang2019racial}. The societal impact of this bias is also huge because the compressed models are commonly used in consumer devices for daily use such as mobile phones and personal assistant devices.


\begin{figure}[t]
\centering
    \includegraphics[width=\columnwidth]{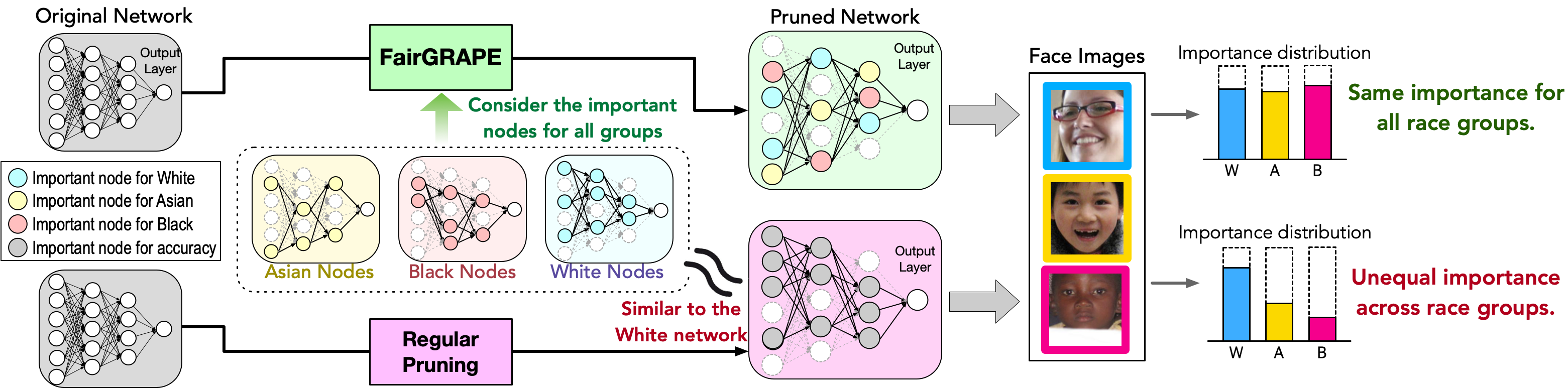}
\caption{Illustration of our proposed model compression method for face attribute classification. Since the compressed model pruned by a regular pruning method shows disparity results over different groups, our proposed pruning method aims to fairly treat all groups by preserving important nodes for sensitive attributes (\eg race, gender) in networks.}
\label{fig:fairgrad}
\end{figure}

To address this critical issue, we propose a novel model pruning method, Fairness-aware GRAdient Pruning mEthod -- \textbf{FairGRAPE}. 
Our method aims at preserving per-group accuracy as well as overall accuracy in classification tasks.  
Figure~\ref{fig:fairgrad} illustrates the fundamental idea of our proposed method. 
Existing pruning methods disregard demographic groupings and prune the nodes with the smallest weights to preserve the model's overall accuracy. However, some nodes may be critical only for a sub-population underrepresented in the dataset and consequently pruned, leading to a biased compressed model. In contrast, our method considers each node's importance to each sub-group separately so that it can retain important features for all groups. 


Specifically, our method computes the group-wise importance of each parameter to get the distribution of the total importance of each group in a model. It then iteratively selects network edges that most closely maintain both the magnitude and share of importance for each group. By selecting such edges, our method equalizes the importance loss across groups, reducing performance disparity. 




To evaluate the effectiveness of our method, we conduct extensive experiments on the face attribute classification tasks where demographic labels are readily available. We use four popular face datasets, FairFace~\cite{karkkainen2019fairface}, UTKFace~\cite{zhifei2017utk}, CelebA~\cite{liu2015CelebA}, and the person subtree in the ImageNet~\cite{yang2020towards}. 
The experimental results show that FairGRAPE not only preserves the overall classification accuracy but also minimizes the performance gap between sub-groups after pruning, compared to other state-of-the-art pruning methods. 
We summarize our contributions as follows.
\begin{itemize}
\item We show that existing pruning methods disproportionately prune important features for different demographic groups, leading to a more considerable accuracy disparity in the compressed model than in the original model.
\item We propose a novel, simple, and generally applicable pruning method that maintains the layer-wise distribution of group importance. 
\item We evaluate our method on four large-scale face datasets compared to four widely used pruning methods. 
\end{itemize}

\section{Related Work}

\subsection{Model Compression via Pruning}
Compression of deep models involves various methods to reduce computation cost without significant loss in model performance. Major categories of compression techniques include Parameter pruning~\cite{frankle2018lottery,han2015weightandconnection}; Parameter quantification \cite{han2015deep}; Lower-rank factorization~\cite{tai2016convolutional_factorization}; knowledge distillation \cite{hinton2015distilling}. In this paper, we focus on examining the first one: parameter pruning, which reduces the number of weights associated with nodes or edges in a network. 


Prior research in pruning has focused on the following aspects: how to maintain certain structural elements of the original model \cite{li2016filterpruning,he2017channel,wang2019corrfilterpruning}, how to rank the importance of individual features~\cite{molchanov2019importancepruning,wang2020grasp,yu2018nisp, lee2018snip, dubey2018coresetpruning, liu2017sliming, molchanov2016pruningresourceinference}, whether pruning should be done at once or across several steps \cite{frankle2018lottery, you2019earlybirdticket}, and how many pruning and retraining iterations are required \cite{blalock2020statepruning,han2015weightandconnection}.

\subsection{Fairness in Computer Vision}
Fairness has received much attention in the recent literature on computer vision and deep learning~\cite{buolamwini2018genderdisparity, wang2019racial,Wang_2019_ICCV_Balanced_dataset,Li_2019_CVPR_Repair,Xu_2021_CVPR_instance_FP,Misra_2016_CVPR,chen2021understanding,Lukasz_2019_Harm_of_demo_bias,Niklas_2020_IEEE_Face_quality_and_bias,yang2022aies,wang2022measuring,jung2022learning,hazirbas2021towards}. The most common goal in these works is to enhance fairness by reducing the \emph{accuracy disparity} of a model between images from different demographic sub-groups. For example, a face attribute classifier may yield a disproportionately higher error rate on images of non-White or females~\cite{buolamwini2018genderdisparity,karkkainen2019fairface}. 
Another line of work has investigated biased or spurious associations in public image datasets and models between different dimensions of sensitive groups and non-protected attributes such as semantic descriptions, facial expressions, and age~\cite{zhao2017men,joo2020gender,zhao2021understanding,chen2021understanding, alvi2018biasremovalembedding}. Our paper focuses on the former: the mitigation of accuracy disparity. 


The cause of demographic bias can be demographically imbalanced datasets and the design choice of learning algorithms or network architectures~\cite{du2020deepfairness,krishnan2020genderbias}. Prior works have found that a face dataset dominated by the White race produces a poor performance for other races, while a face dataset with balanced group distribution, from either real or synthesized data, can enhance fairness~\cite{karkkainen2019fairface,GEORGOPOULOS_IVC_2020_KANFace,frid2018ganfairness, Wang_2019_ICCV_Balanced_dataset,Gwilliam_2021_ICCV,yang2022explaining}. Algorithmic bias can be mitigated through either explicit fairness constraints~\cite{zafar2017fairnessconstraints,Matth_2019_ICML_Fairness_constraint}, matching learned representations to a target distribution or group-wise characteristics~\cite{das2018multitaskCNN,ryu2017inclusivefacenet,schumann2019transferfairness}, or adversarial mitigation and decoupling to disconnect representation and sensitive groups that attempts to decorrelate sensitive attributes and model outputs~\cite{Ramaswamy_2021_CVPR_latent_fair,alvi2018biasremovalembedding, Hyojin2020debiased,Wang_2020_CVPR,lee2021learning,dwork2018decoupled}. Our method estimates the importance of each connection weight toward each sub-group and maintains between-group ratios in pruning. 

\subsection{Fairness in Model Compression}
Only a few studies have been concerned with fairness in the compression of deep models. 
\cite{hooker2019compressed} reported that pruned models tend to forget specific subsets of data. It is examined in \cite{hooker2020bias, stoychev2022compressionEffect} that pruning can impact demographic sub-groups disproportionately in face attribute classification and expression recognition. 
Another recent work \cite{blakeney2021simon} showed that knowledge distillation could reduce bias in pruned models. All these studies focus on measuring pruning-induced biases between output categories. To the best of our knowledge, our paper is the first to separate the pruning impact on output classes and sensitive groups and propose a pruning algorithm to mitigate biases in both dimensions.

\section{Fairness-aware GRAdient Pruning mEthod}

\subsection{Problem Statement and Objective}
\label{sec:objective}

Consider a neural network $\bm{f}$ parameterized by $\bm{\theta} \in \mathbb{R}^m$ and a dataset $D = \{(\bm{x_i}, \bm{y_i}, k_i)\}^{n}_{i=1}$, where $\bm{x_i} \in \mathcal{X}$ is an input vector, $\bm{y_i} \in \mathcal{Y}$ is a target output, and $k_i \in K$ is a sensitive attribute. The goal of network pruning is to find the following parameter set:

\begin{align}
    \bm{\theta'} = \argmin_{\bm{\theta}}L(D,\bm{\theta}) = \argmin_{\bm{\theta}}\frac{1}{n} \sum_{i=1}^n \ell(f(\bm{\theta}, \bm{x_i}), \bm{y_i}), \text{ subject to }||\bm{\theta}||_0 \leq c \cdot m
\end{align}

Here $\ell(\cdot)$ denotes a loss function, and $c \in (0,1)$ is the desired sparsity level. 



We further examine the network's performance on different subsets of $D$. Let $D_k = \{(\bm{x_i}, \bm{y_i}, k_i) | k_i = k \}$ denote the subset of instances from a sensitive group $k \in K$. Given a performance metric $A(D_{k};\bm{\theta})$, the difference in performance on $D_k$ between the full model $\bm{f}$ and a compressed model $\bm{f'}$ is:
\begin{align}
     \Delta A(D_{k};\bm{\theta'}) & = A(D_{k};\bm{\theta'})-A(D_{k};\bm{\theta})
\end{align}

The mean of all group-wise performance differences is:
\begin{align}
     \Delta \Bar{A}(D; \bm{\theta'}) & = \frac{1}{|K|} \sum_{k \in K} \Delta A(D_{k};\bm{\theta'})
\end{align}

Our goal is to minimize the variance of performance differences in a pruned model. This task can be formulated as finding the following $\bm{\theta^*}$:
\begin{align}
    \bm{\theta^*} = 
     \argmin_{\bm{\theta'}} [Var(\Delta A(D_k;\bm{\theta'}))] & = \argmin_{\bm{\theta'}} [ \space \frac{1}{|K|}  \sum_{k \in K}(\Delta A(D_k;\bm{\theta'})- \Delta \Bar{A}(D; \bm{\theta'}))^2]
\end{align}



Note that the actual task of the model determines the choices of the performance metric $A(\bm{\theta};D_{k})$. This paper focuses on classification tasks and thus uses accuracy, false positive rate~(FPR), and false negative rate~(FNR) as performance metrics. The output space $\mathcal{Y}$ and sensitive groups $K$ can be either overlapping or disjoint, and this paper examines both cases.

\begin{figure*}[!t]
\centering
\includegraphics[width=0.98\textwidth]{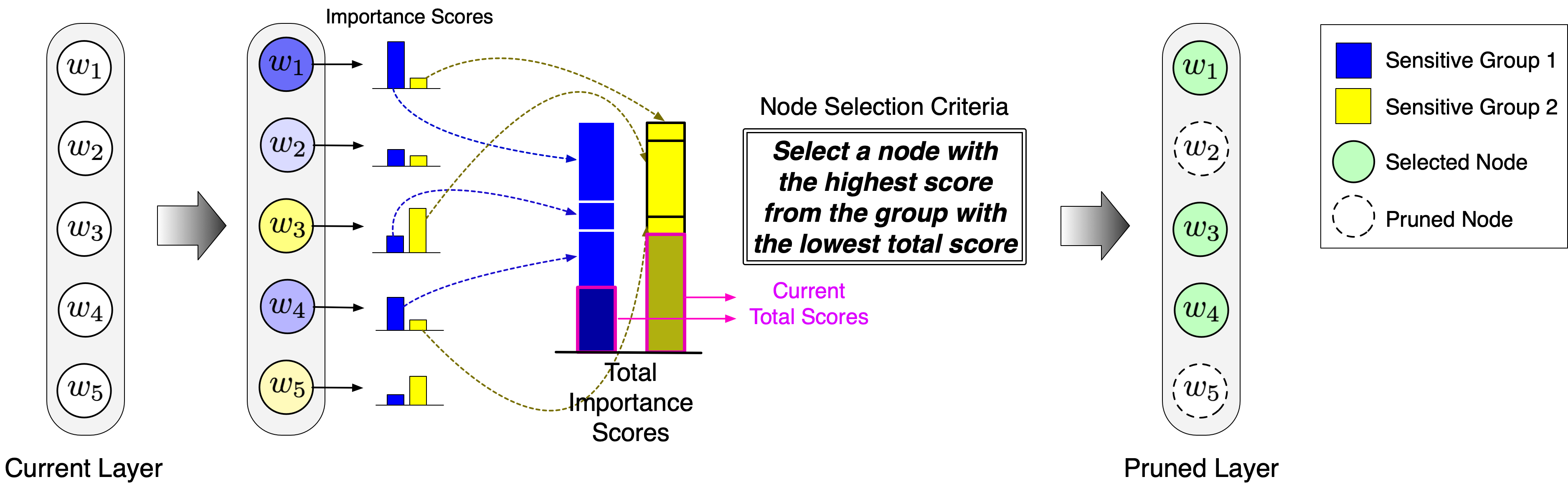} 
\caption{Illustration of the proposed node selection method. FairGRAPE first computes the importance score of each individual weight for all groups layer-wise. Based on the total scores from the current layer, FairGRAPE selects a node with the highest score from the group with the
greatest loss in importance score to minimize the variance of performance changes.}
\label{fig:framework}
\end{figure*}

\subsection{FairGRAPE: Fairness-aware  Gradient Pruning Method}

The common idea behind model pruning methods is estimating the importance of edges and pruning less important ones. While the existing methods focus on measuring the importance to the whole dataset, our method aims to preserve important weights for each sensitive group to mitigate biases.

To this end,we propose to compute the group-wise importance score of each weight with respect to each sensitive group,
and then use a greedy algorithm to select weights based on the scores. 
At each step, the method compares the current ratio of importance scores with the target ratio (\ie, the ratio in the model before the current pruning step).
Then the group that has the largest difference will be selected, and the method adds one weight with the highest importance for the selected group to the selected network. Once the desired number of weights is selected, the remaining weights are pruned.
FairGRAPE compresses all layers of the model with this node selection process, which is illustrated in Figure~\ref{fig:framework}. 


\subsubsection{Group-wise Importance}

Let $w \in \bm{\theta'}$ denote a parameter in $f'$ and $L(D_k, \bm{\theta'})$ denote the loss on sensitive group $k$. The gradient of $L(D_k, \bm{\theta'})$ with respect to $w$ is $g_{w} = \frac{\partial L(D_k, \bm{\theta'})}{\partial w}$.
Then the importance of $w$ with respect to group $k$ and the total model importance score for a group $k$ are:
\begin{align}
& I_{k,w} = (L(D_k, \bm{\theta'})-L(D_k, \bm{\theta'}|w=0))^2
& I_{k}  =  \sum_{w \in \bm{\theta'}} {I_{k,w}}
\label{eq:importance}
\end{align}

Computing the importance defined in equation \ref{eq:importance} requires evaluating a different network for every parameter, which is often impractical. Alternatively, $I_{w,k}$ could be approximated by its first-order Taylor expansion, as explained in~\cite{molchanov2019importancepruning}:
\begin{align}
    I_{k,w} = (g_{w}w)^2
\end{align}




\subsubsection{Maintaining Share of Importance}


Based on the group importance scores, we compute the share of the importance of group $k$ as follows:
\begin{align}
& P_k =  \frac{I_k}{\sum_{K}{I_k}}
\end{align}

The share of importance in the original model $\bm{f}$ is used as a target. In the pruned model $\bm{f'}$ with parameter set $\bm{\theta'}$, the percentage change in the importance score compared to full model $\bm{f}$ is: 
\begin{align}
& \Delta P_{k,\bm{f'}} = \frac{P_{k,\bm{\theta'}} - P_{k,\bm{\theta}}}{P_{k,\bm{\theta}}}
\end{align}

As weights are pruned, the importance scores for each group would inevitably decrease. However, the disparate loss of importance across groups leads to an imbalanced loss in classification performance. Thus, we apply a layer-wise greedy algorithm to select the parameters that minimize the difference of $\Delta P_{k, \bm{f`}}$ between the sensitive groups, as explained in Algorithm~\ref{alg:GDP}.

\begin{algorithm}[t]
\caption{FairGRAPE}\label{alg:GDP}
{\small 
\begin{algorithmic}[1]
\State $c \gets$ desired sparsity
\State $r \gets \%$ of parameters to prune per iteration
\State $Iters\gets \lceil log_{r}(c) \rceil$; $i \gets 0$
\State $f \gets $ pre-pruning network with parameter set $\bm{\theta}$
\While{$i < Iters$}
\For{$\bm{\theta}_{layer}$ \textbf{in} $\bm{\theta}$}
\State $\bm{\theta'}_{layer} \gets \emptyset$ 
\While{$|\bm{\theta'}_{layer}| <  ||\bm{\theta}_{layer}||_0 \times (1 - r)$}
\State $\Tilde{k} \gets \min_{k}(\Delta P_{k,\bm{\theta'}_{layer}})$ \Comment{{\footnotesize Find the greatest importance loss}} \label{lst:findk}
\State $\Tilde{w} \gets \argmax_{w \in \bm{\theta'}_{layer}}( I_{\Tilde{k},w})$ 
\Comment{{\footnotesize Find the highest importance for $\Tilde{k}$ }} \label{lst:findw}
\State $\bm{\theta'}_{layer} \gets \bm{\theta'}_{layer} \bigcup   \{\Tilde{w}\}$
\State $\Delta P_{k,\bm{\theta'}_{layer}} \gets \frac{P_{k,\bm{\theta'}_{layer}} - P_{k,\bm{\theta}_{layer}}}{P_{k, \bm{\theta}_{layer}}}, \forall k \in K$ \Comment{{\footnotesize Update importance losses}} \label{lst:updatep}
\EndWhile
\State $\bm{\theta}_{prune} \gets \{w | w \in  \bm{\theta}_{layer} \land w \notin \bm{\theta'}_{layer} \}$
\State $\bm{\theta}_{prune} \gets \bm{0}$ \Comment{{\footnotesize Prune weights that are not selected}}
\EndFor
\State $f \gets$ Train $(\bm{\theta}; D)$; $i \gets i+1$ 
\EndWhile
\end{algorithmic}
}
\end{algorithm}

FairGRAPE iteratively prunes and fine-tunes a network: given a desired sparsity $c$ and a step size $r$, the percentage of remaining weights to be pruned at each iteration, the total number of iterations is $Iters = \lceil (log_{r}(c) \rceil$.  In each iteration, the network is pruned layer by layer for all layers with weight attributes (\eg convolutional and linear layers). Before pruning a layer,  $P_{k,\bm{\theta}}$ for each group $k$ is calculated with all unpruned parameters $\bm{\theta}$. At the very beginning of the algorithm, $\bm{\theta'}$ has not included any weights yet, and all group importance values are $0$. So $P_{k, \bm{\theta'}}$ are initialized to $1/|K|$. Then weights in $\bm{\theta}$ are added to $\bm{\theta'}$ one at a time. Before each selection, the sensitive group $k$ with the minimum $\Delta P_{k, \bm{\theta'}}$ is identified, as shown in line~\ref{lst:findk} of Algorithm~\ref{alg:GDP}. Then the weight that has the highest importance score for group $k$ is added to the set of selected weights to minimize $P_{k,\bm{\theta'}}$ (line~\ref{lst:findw}). $P_{k, \bm{\theta'}}$ and $\Delta P_{k, \bm{\theta'}}$ are updated for all groups $k \in K$ (line~\ref{lst:updatep}). The selection for weights continues until $(1-r)$\% of weights are selected. The weights not selected are removed by setting them to zero and thus no longer considered in further iterations. Then FairGRAPE proceeds to the next layer. Once all layers are pruned, the network is retrained for a fixed number of $e$ epochs to adjust the weights to its current structure. Then the next iteration begins.

\section{Experiments}

\subsection{Datasets}


To evaluate our proposed FairGRAPE, we conducted extensive experiments with four face image datasets, including FairFace~\cite{karkkainen2019fairface}, UTKFace~\cite{zhifei2017utk}, CelebA~\cite{liu2015CelebA}, and the person subtree of ImageNet~\cite{yang2020towards}.
Table~\ref{tab:datasets} shows the distributions of races and genders in all datasets. Images are fairly distributed across the seven race groups in the FairFace, while the white race is dominant in the UTKFace. This allows us to validate that the effect of our method remains consistent with the presence of data bias. In UTKFace, only one ``\textit{Asian}'' contains both Asian and Southeast Asian faces. We excluded the ``\textit{Other}'' category in UTKFace due to its ambiguity. Race/ethnicity information is not provided in CelebA and ImageNet. 
FairFace, UTKFace, and CelebA provide annotations for binary genders. While the ImageNet person subtree contains three gender classes: Male, Female, and Unsure (non-binary), we only use ImageNet samples with binary genders to stay consistent with other datasets. Following the practice in~\cite{yang2020towards}, Imagenet samples that are from "unsafe" categories or have imageability scores $\leq$ 4 are also excluded.


\begin{table}[t!]
    \centering
    \resizebox{\linewidth}{!}{
    \begin{tabular}{c||c|cccC{1cm}C{1.4cm}cC{1.2cm}|cc|c}
        Dataset & Images & White & Black& Hispanic & East Asian & Southeast Asian & Indian & Middle Eastern& Male & Female& Categories \\
         \hline
        FairFace\cite{karkkainen2019fairface} & 97,698 & 18,612 & 13,789 & 14,990& 13,837 & 12,210& 13,835& 10,425& 51,778& 45,920 & - \\
        UTKFace\cite{zhifei2017utk} & 22,013 & 10,078 & 4,526&-&3,434&-&3,975&-&11,631&10,382&-  \\
        CelebA\cite{liu2015CelebA} &202,599&-&-&-&-&-&-&-&84,434& 118,165&39 \\
        ImageNet(Person)\cite{yang2020towards} & 10,215 & - & -& -& -& -& -& - & 6,590 & 3,625 & 103\\
        \hline
    \end{tabular}}
    \caption{Demographic composition of datasets.} 
    \label{tab:datasets}
\end{table}


\subsection{Experiment Settings}
\noindent\textbf{Network architectures}: To ensure our method applies to different architectures, we use two popular deep networks: ResNet-34 \cite{he2016deepResnet} and MobileNet-V2 \cite{howard2017mobilenets}.
ResNet is widely applied for classification tasks, and the MobileNet is a compact network commonly used by mobile devices. All models are pre-trained on ImageNet~\cite{deng2009imagenet}.

\noindent\textbf{Hyperparameters}: 
We use a cross-entropy loss function with the ADAM optimizer for all training.   
All accuracy scores, overall and group-wise, are averaged across three trials to control for randomness in training. 
For iterative pruning methods, we retrain five epochs after each pruning iteration. Step size $r$ = 0.9 on FairFace, CelebA and Imagenet and $r$ = 0.975 on UTKFace. The training/validation/testing percentage is 80\%/10\%/10\% in each dataset. 

\subsection{Baseline Methods}
We deploy the following four baseline methods: \textbf{Single-Shot Network Pruning (SNIP)}~\cite{lee2018snip}: calculates the connection sensitivity of edges by back-propagating on one mini-batch and prunes the edges with low sensitivity. 
\textbf{Weight Selection (WS)}~\cite{han2015deep}: prunes the weights with magnitudes below a threshold in a trained model. It is the most commonly used in mobile applications\cite{nan2019deeponmobile}. \textbf{Lottery Ticket Identification (Lottery)}~\cite{frankle2018lottery}: records the initial state of the network; resets the model to its initial state after each pruning iteration. \textbf{Gradient Signal Preservation (GraSP)}~\cite{wang2020grasp}: removes the parameters with low Hessian-gradient scores to maximize gradient signal in the pruned model.



\section{Results}
To evaluate the effectiveness of our method, we conduct extensive experiments on three different settings, including (\cref{sec:exp1}) gender and race classification tasks, (\cref{sec:exp2}) non-sensitive attribute classification tasks, and (\cref{sec:exp3}) model pruning based on unsupervised clustering.
We also perform more in-depth analysis, including (\cref{sec:analysis1}) ablation studies, (\cref{sec:analysis2}) different sparsity levels, (\cref{sec:minority}) pruning on minority faces, 
and (\cref{sec:analysis3}) difference in importance score and structure, to understand the importance of components in FairGRAPE.

\subsection{Gender and Race Classification} \label{sec:exp1}

\begin{table}[!t]
\renewcommand{\arraystretch}{1.0}%
\centering
\resizebox{1\linewidth}{!}{
\begin{tabular}{c|c|| c | cc|cc || c | ccccccc|cc} 
 \multirow{2}{*}{Task} & \multirow{2}{*}{Method} & \multicolumn{3}{c}{Accuracy}  & \multicolumn{2}{c}{Bias }  \vline  &   \multicolumn{8}{c}{Accuracy}  & \multicolumn{2}{c}{Bias}   \\
\cline{3-17}
  & & All & Male & Female & $\rho(A)$ & $\rho(\Delta) $ & All & White  &  Black & Hisp & E-A & SE-A & Indian & ME &  $\rho(A)$ & $\rho(\Delta)$ \\
 \hline
 \hline
 \multirow{6}{*}{\begin{tabular}{c}
      FairFace,  \\
      Gender\\
 \end{tabular}} & No-pruning & 94.6 & 94.7 & 94.5 & 0.14 & - & 94.6 & 94.6 & 90.5 & 95.9 & 94.7 & 94.4 & 96.3 & 95.6 & 1.93 & - \\
    \cline{2-17}
    & Lottery & 85.8 & 86.4 & 85.2 & 0.80 & 0.65 & 85.8 & 85.1 & 80.8 & 88.4 & 84.0 & 85.5 & 88.1 & 89.6 & 3.01 & 1.55 \\
    &  SNIP & 90.4 & 91.0 & 89.9 & 0.78 & 0.63 & 90.4 & 91.0 & 85.2 & 92.6 & 90.0 & 90.5 & 91.3 & 92.6 & 2.53 & 0.93 \\ 
    &  WS & 83.8 & 84.3 & 83.4 & 0.62 & 0.47 & 83.9 & 82.9 & 78.9 & 87.2 & 82.2 & 82.2 & 86.2 & 88.3 & 3.32 & 2.00 \\ 
    &  GraSP & 87.9 & 88.4 & 87.4 & 0.75 & 0.60 & 87.9 & 87.5 & 83.1 & 89.6 & 87.5 & 88.0 & 89.4 & 90.9 & 2.49 & 0.93 \\ 
    &  \textbf{FairGRAPE} & \textbf{91.1} & 91.3 & 91.0 & \textbf{0.20} & \textbf{0.05} & \textbf{90.5} & 90.4 & 85.4 & 92.3 & 90.1 & 90.5 & 91.9 & 92.8 & \textbf{2.47} & \textbf{0.77}  \\ 
  \hline 
  \multirow{6}{*}{\begin{tabular}{c}
      FairFace,  \\
      Race\\ 
 \end{tabular}} & No-pruning  & 72.0 & 71.2 & 72.9 & 1.23 & - & 72.0 & 73.9 & 83.2 & 59.6 & 77.6 & 66.9 & 75.4 & 66.2 & 8.02 & - \\ 
    \cline{2-17}
    &  Lottery & 57.1 & 55.3 & 59.1 & 2.64 & 1.42 & 57.1 & 69.7 & 78.8 & 33.0 & 74.1 & 43.5 & 61.7 & 30.4 & 20.0 & 12.9 \\
    &  SNIP & 62.3 & 60.4 & 64.3 & 2.78 & 1.55 & 62.3 & 74.1 & 80.8 & 44.5 & 73.7 & 53.7 & 66.0 & 34.8 & 17.1 & 10.7 \\ 
    &  WS & 47.9 & 47.3 & 48.5 & \textbf{0.86} & \textbf{0.36} & 47.9 & 64.7 & 77.9 & 8.61 & 78.3 & 31.1 & 37.8 & 30.0 & 26.9 & 19.9 \\ 
    &  GraSP  & 57.9 & 56.0 & 60.1 & 2.88 & 1.55 & 57.9 & 69.6 & 77.3 & 38.6 & 72.0 & 47.0 & 62.1 & 30.7 & 18.0 & 11.3 \\  
    &  \textbf{FairGRAPE} & \textbf{66.8} & 65.3 & 68.6 & 2.35 & 1.12 & \textbf{65.1} & 72.2 & 80.3 & 47.5 & 75.8 & 56.3 & 70.2 & 48.6 & \textbf{13.4} & \textbf{6.13} \\
  \hline 
  \multirow{6}{*}{\begin{tabular}{c}
      UTKFace,  \\
      Gender\\
 \end{tabular}} & No-pruning & 93.5 & 92.4 & 94.8 & 1.68 & - & 93.5 & 94.1 & - & 95.1 & - & 89.6 & - & 93.7 & 2.45 & - \\  
    \cline{2-17}
    &  Lottery & 83.5 & 83.7 & 83.3 & 0.34 & 2.01 & 83.5 & 84.7 & - & 85.8 & - & 75.0 & - & 85.2 & 5.15 & 2.79 \\ 
    &  SNIP & 91.0 & 91.3 & 90.6 & 0.45 & 2.19  & 91.0 & 91.9 & - & 93.0 & - & 86.0 & - & 90.9 & 3.08 & 0.67 \\
   & WS & 81.9 & 81.4 & 82.6 & 0.89 & 1.79 & 81.9 & 82.1 & - & 84.9 & - & 77.2 & - & 82.4 & 3.20 & 0.92 \\ 
    &  GraSP & 86.8 & 88.5 & 84.9 & 2.51 & 4.20 & 86.8 & 86.7 & - & 89.8 & - & 81.4 & - & 88.3 & 3.66 & 1.43 \\ 
    &  \textbf{FairGRAPE} & \textbf{92.2} & 92.0 & 92.5 & \textbf{0.31} & \textbf{1.36} & \textbf{91.9} & 92.7 & - & 94.0 & - & 87.9 & - & 91.3 & \textbf{2.61} & \textbf{0.56} \\
  \hline 
  \multirow{6}{*}{\begin{tabular}{c}
      UTKFace,  \\
      Race\\ 
 \end{tabular}}& No-pruning & 90.8 & 90.6 & 90.9 & 0.24 & - & 90.8 & 92.2 & - & 92.5 & - & 93.3 & - & 83.3 & 4.69 & - \\
    \cline{2-17}
    &  Lottery & 71.7 & 69.4 & 74.2 & 3.41 & 3.17 & 71.7 & 83.8 & - & 80.3 & - & 61.0 & - & 42.7 & 19.0 & 15.6 \\ 
    &  SNIP & 86.8 & 85.7 & 88.0 & 1.64 & 1.40 & 86.8 & 91.6 & - & 92.5 & - & 85.8 & - & 70.1 & 10.4 & 6.28 \\
    &  WS & 70.7 & 68.3 & 73.5 & 3.68 & 3.41 & 70.7 & 82.7 & - & 80.8 & - & 59.2 & - & 41.4 & 19.6 & 16.2 \\
    &  GraSP & 77.7 & 76.4 & 79.1 & 1.94 & 1.70 & 77.7 & 86.1 & - & 83.3 & - & 72.2 & - & 56.4 & 13.5 & 9.81 \\
    & \textbf{FairGRAPE} & \textbf{88.7} & 88.2 & 89.3 & \textbf{0.78} & \textbf{0.54} & \textbf{88.5} & 90.6 & - & 92.2 & - & 88.9 & - & 79.0 & \textbf{5.93} & \textbf{2.04} \\ 
  \hline    

\end{tabular}}
\caption{The group-wise accuracy and biases in gender or race classification tasks. Hisp, E-A, SE-A and ME stand for Hispanic, East Asian, Southeast Asian and Middle Eastern. FairFace experiments are conducted on ResNet-34 pruned at 99\% sparsity, UTKFace experiment on MobileNet-V2 pruned at 90\% sparsity. $\rho (A)$ and $\rho (\Delta)$ are the standard deviation of accuracy and accuracy loss across sensitive groups, respectively. } 
\label{tab:FF_UTK_results}
\end{table}

We first perform experiments to verify bias mitigation in classifying sensitive attributes. 
Table~\ref{tab:FF_UTK_results} shows classification accuracy and biases on FairFace and UTKFace datasets where we compress the ResNet-34 and MobileNet-V2, respectively. The column `Task' indicates the dataset and classification task. We report overall classification accuracy, accuracy by sensitive groups, and variances in accuracy degradation. 
FairGRAPE consistently produces a substantially higher accuracy, lower differences in accuracy, and lower variance in performance degradation than the baseline methods.
For example, SNIP sometimes produces accuracy scores close to our method, but it has a remarkably larger accuracy variance than FairGRAPE, which implies the potential biases caused by model pruning. In the only cases of FairFace, WS produced a model with a smaller race classification accuracy gap between male and female images, but at the cost of drastically worsened accuracy for both groups.
These results suggest that our proposed method successfully equalizes the impact of pruning on the sensitive groups regardless of the classification task, thus achieving a better trade-off between fairness and overall accuracy.

Furthermore, FairGRAPE shows solid performances in all
settings with different architectures and datasets (balanced or imbalanced), proving the proposed method’s robustness. See supplementary material for results when we jointly control race and gender groups. 

\begin{figure*}[!t]
\centering
    \includegraphics[width=\linewidth, height=0.3\linewidth]{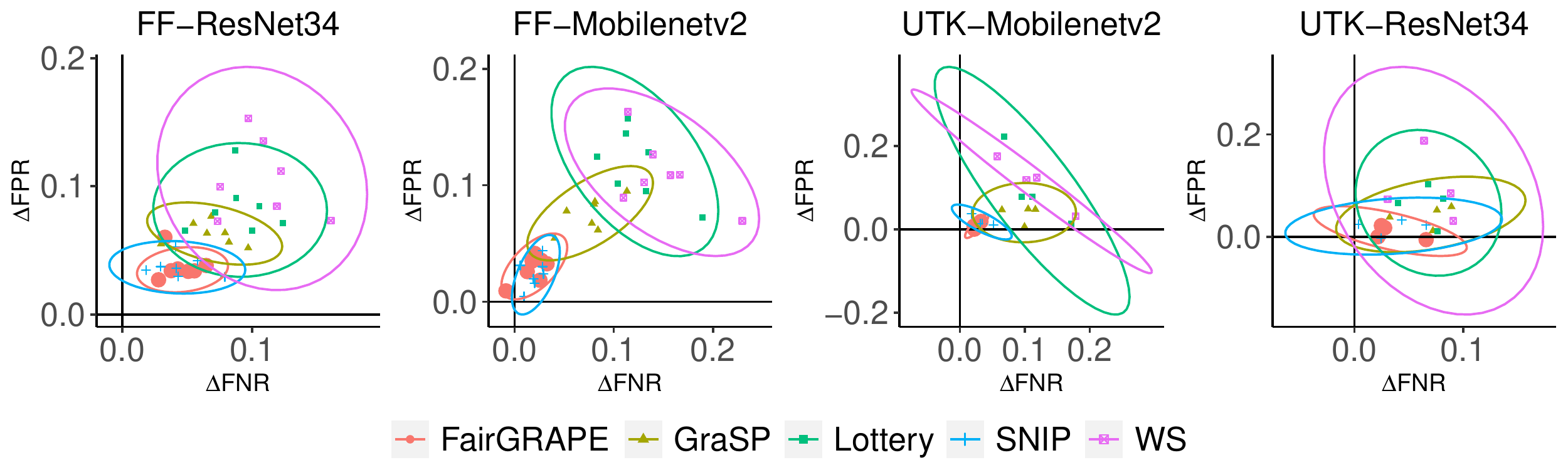} 
     \caption{Normalized FNR/FPR changes in race classification. Sparsity levels are 99\% for ResNet-34 and 90\% for MobileNet-V2. Each data point represents the mean value of a race. The ellipses are created by estimating 95\% confidence ellipses, assuming multivariate $t$-distribution of points produced by each method. } 
    \label{fig:delta_FR}
\end{figure*}

We next visualize the proportion changes of False Negative Rates(FNRs)/False Positive Rate(FPRs) from the full model after pruning by FairGRAPE and other baseline methods in Figure~\ref{fig:delta_FR}.
Each point in the plot represents normalized FNR and FPR change of a specific race group in the model produced by one of the pruning methods, and the ellipses are created by estimating a 95\% confidence region of data points. 
The results reveal that the proposed FairGRAPE produces data points closer to the origin than the other data points generated by the baseline methods.
More importantly, FairGRAPE creates the smallest ellipse, which demonstrates that performance changes for each group are close to each other. Thus the distribution of induced bias across sensitive groups is fair. 

\subsection{Non-Sensitive Attribute Classification} \label{sec:exp2}


To evaluate the performance of FairGRAPE in more practical cases where output classes and sensitive groups are disjoint, we experiment with classification on CelebA and ImageNet datasets. CelebA contains the 39 non-sensitive categories of facial attributes such as eyeglasses, makeup, and lipsticks. We code each of these categories as a binary classification task. For the ImageNet experiment, we use the modified person subtree, which contains 10,215 images in 103 distinct classes (\eg, basketball player, rapper) with gender labels~\cite{yang2020towards}.
We train the models to classify the class to which a given image belongs.
Note that we use the ResNet-34 network at 50\% sparsity for the ImageNet experiments and the MobileNet-V2 network at 90\% sparsity for the CelebA experiments. 

Table~\ref{tab:CelebA_result} shows the overall accuracy, accuracy of each gender, and the standard deviation of accuracy change. 
FairGRAPE achieves the highest accuracy on ImageNet. Although GraSP has a smaller accuracy gender gap than our method, its overall accuracy and variance of performance degradation are drastically worse. In the CelebA experiment, FairGRAPE has a significantly lower variance in accuracy change than other methods while achieving the highest accuracy. The results demonstrate that FairGRAPE performs well on sensitive attribute classification tasks and non-sensitive attributes, thus widely applicable in various applications.

\begin{table}[!t]
\renewcommand{\arraystretch}{1.0}%
\centering
\resizebox{0.8\linewidth}{!}{
\begin{tabular}{c|c|c|c|c|cc|cc} 
  \multirow{2}{*}{Dataset} &  \multirow{2}{*}{Task} &  \multirow{2}{*}{Group} & \multirow{2}{*}{Methods} &  \multicolumn{3}{c}{Accuracy} \vline & \multicolumn{2}{c}{Bias}   \\ \cline{5-9}
  & & & & All & Male & Female & Diff & $\rho (\Delta)$ \\
 \hline \hline
    \multirow{6}{*}{ImageNet} & \multirow{6}{*}{\thead{Person\\ Subtree \\ (103 classes)}} & \multirow{6}{*}{Gender} & No-Pruning & 50.25 & 53.03 & 45.60 & 7.43 & -   \\ 
    \cline{4-9}
    & & & Lottery & 50.85 & 54.03 & 45.98 & 8.05 & 2.55 \\
    & & & SNIP & 47.85 & 50.89 & 42.76 & 8.13 & 0.49 \\ 
    & & & WS & 51.11 & 54.06 & 46.16 & 7.90 & 0.33 \\
    & & & GraSP & 15.36 & 17.03 & 12.57 & 4.47 & 2.10 \\
    & & & \textbf{FairGRAPE}& \textbf{51.12} & 54.01 & 46.16 & \textbf{7.85} &\textbf{0.30} \\ 
 \hline  \hline
 \multirow{6}{*}{CelebA} & \multirow{6}{*}{\thead{Non-sensitive \\ Facial \\Attributes \\ (39 classes)}} & \multirow{6}{*}{Gender} & No-Pruning & 91.81 & 91.76 & 91.86 & 0.11 & -  \\ 
 \cline{4-9}
    & &   & Lottery & 89.31 & 88.99 & 89.54 & 0.55 & 0.32 \\
    & &   & SNIP & 90.29 & 90.05 & 90.46 & 0.41 & 0.21 \\ 
    & &   & WS & 88.57 & 88.15 & 88.87 & 0.72 & 0.43 \\ 
    & &   & GraSP & 89.40 & 89.08 & 89.63 & 0.55 & 0.32   \\
    & &   & \textbf{FairGRAPE} & \textbf{90.90} & 90.74 & 91.01 & \textbf{0.27} & \textbf{0.11} \\
 \hline 
\end{tabular}} 
\caption{The average accuracy and biases in person category classification and facial attributes classification on ResNet-34 at 50\% sparsity and MobileNet-V2 network at 90\% sparsity, respectively. $\rho (\Delta)$ is the standard deviation of accuracy loss across genders.  } 
\label{tab:CelebA_result}
\end{table}

\subsection{Unsupervised Learning for Group Aware in Model Pruning} \label{sec:exp3}

In practice, labels for sensitive attributes may not always be available. Therefore, we further examine the performance of our method on a dataset without demographic group labels through unsupervised group discovery. 


Table~\ref{tab:pseudoGroup} shows the accuracy and bias of experiments on the FairFace dataset. In this test, FairGRAPE conducts pruning by calculating the importance score of parameters for clusters learned from unsupervised learning as sensitive groups. Then we evaluate accuracy and bias with actual race labels. We labeled the seven clusters using K-means clustering on image embedding generated by the ResNet-34 network pre-trained on Imagenet. 
While all baseline methods have low accuracy and large variance of accuracy as they do not consider the sensitive groups, the FairGRAPE method consistently results in the lowest performance variance, suggesting that our proposed method has the potential to compress the model while reducing biases even in the absence of sensitive attribute information. The K-means algorithm's simplicity further reinforced our method's generalizability when the precise group partitioning is complex or noisy. 

\begin{table}[!t]
\renewcommand{\arraystretch}{1.0}%
\centering
\resizebox{0.8\linewidth}{!}{
\begin{tabular}{c|c||c|ccccccc | cc} 
  \multirow{2}{*}{Task} & \multirow{2}{*}{Methods}  & \multicolumn{8}{c}{Accuracy} \vline & \multicolumn{2}{c}{Bias} \\
\cline{3-12}
  &  & All & White  &  Black & Hisp & E-A & SE-A & Indian & ME & $\rho(A)$ & $\rho (\Delta)$ \\
 \hline
 \hline
 \multirow{6}{*}{\begin{tabular}{c}
      FairFace,  \\
      Race\\ 
 \end{tabular}} & No-Pruning & 72.0 & 73.9 & 83.2 & 59.6 & 77.6 & 66.9 & 75.5 & 66.2 & 8.03 & - \\ 
    \cline{2-12}
    & Lottery& 57.1 & 69.7 & 78.8 & 33.0 & 74.1 & 43.5 & 61.7 & 30.4 & 20.0 & 12.9 \\  
    & SNIP & 62.3 & 74.1 & 80.8 & 44.5 & 73.7 & 53.7 & 66.0 & 34.8 & 17.1 & 10.7 \\ 
    & WS  & 47.9 & 64.7 & 78.0 & 8.6 & 78.3 & 31.1 & 37.8 & 30.0 & 26.9 & 19.9 \\ 
    & GraSP & 57.9 & 69.6 & 77.3 & 38.6 & 72.0 & 47.0 & 62.1 & 30.7 & 18.0 & 11.3 \\
    & \textbf{FairGRAPE} & \textbf{63.5} & 69.7 & 80.4 & 49.2 & 74.7 & 53.7 & 68.1 & 42.8 & \textbf{14.1} & \textbf{7.40} \\ 
  \hline 
\end{tabular}} 
\caption{The average accuracy and biases in race classification, where FairGRAPE pruning is performed based on groups clustered by unsupervised learning. Hisp, E-A, SE-A and ME stand for Hispanic, East Asian, Southeast Asian and Middle Eastern.  $\rho (A)$ and $\rho (\Delta)$ are the standard deviation of accuracy and accuracy loss across sensitive races.}
\label{tab:pseudoGroup}
\end{table}

\subsection{Ablation Studies: Group Importance and Iterative Pruning} \label{sec:analysis1}

\begin{table}[!t]
\renewcommand{\arraystretch}{1.0}%
\centering
\resizebox{0.85\linewidth}{!}{
\begin{tabular}{C{1.7cm}|C{3cm}|C{1.7cm}|c|cc|cc} 
 Group & Iterative & \% Training & \multicolumn{3}{c}{Accuracy} \vline & \multicolumn{2}{c}{Bias}  \\ 
 \cline{4-8}
  Importance &  retraining (\# iterations/$r$) &  Images & All & Female & Male & Diff & $\rho (\Delta)$  \\ \hline \hline       
        \cmark & \cmark \:(22/0.1) & 20\% & 90.90 & 91.01 & 90.74 & 0.27 & 0.11 \\ 
        \cmark & \cmark \:(22/0.1) & 100\% & 90.66 & 90.81 & 90.45 & 0.36 & 0.18  \\  
        \cmark & \cmark \:(22/0.1) & 50\% & 90.72 & 90.84 & 90.54 & 0.30 & 0.13 \\
        \cmark & \cmark \:(22/0.1) & 10\% & 90.84 & 90.97 & 90.67 & 0.30 & 0.13 \\ 
        \cmark & \cmark \:(16/0.2) & 20\% & 90.49 & 90.62 & 90.32 & 0.30 & 0.20 \\
        \cmark & \cmark \:(3/0.5) & 20\% & 90.34 & 90.52 & 90.10 & 0.42 & 0.22  \\ 
        \cmark & \xmark \:(1/0.9) & 20\% & 89.26 & 89.51 & 88.92 & 0.41 & 0.34\\ 
        \xmark & \cmark \:(22/0.1) & - & 89.31 & 89.54 & 88.99 & 0.45 & 0.31  \\ 
        \xmark & \xmark \:(1/0.9) & - & 88.57 & 88.86 & 88.17 & 0.69 & 0.42  \\ 
 \hline 
\end{tabular}} 
\caption{The accuracy and biases under different pruning settings. The MobileNet-V2 networks are trained on CelebA attributes classification tasks and pruned at 90\% sparsity. \# iter is the number of pruning iterations, determined by the pruning step $r$ which is the proportion of remain edges removed during each iteration. \% training images represents the percentage of training images included in calculation of group importance scores. $\rho (\Delta)$ is the standard deviation of accuracy loss across genders. 
}
\label{tab:ablation}
\end{table}

Table~\ref{tab:ablation} shows the performance of FairGRAPE with different group importance and iterative retraining settings.
We first find that group importance is the essential component in our proposed method. The baseline method, which does not use both group importance and iterative retraining, has remarkably lower accuracy, gender gap, and variance of accuracy changes than our method, which utilizes both components. As the pruning step r at each iteration increased, the accuracy decreased, and the bias increased gradually.  

More specifically, the model suffers from an obvious performance drop and bias increase when $r$ increased from 0.1 to 0.2. This result agrees with previous findings~\cite{frankle2018lottery} that iterative pruning improves performance. 

 
Finally, we examine the percentage of training images used in importance calculation. FairGRAPE calculates group-wise importance score $I_{w,k} = (g_{w}w)^2$ for each weight $w$, where $g_w$ is calculated with respect to average loss across selected mini-batches of the training set. It has been found that the proportion of training images used in the calculation process affects pruning speed and accuracy~\cite{molchanov2019importancepruning}. We compared the performance using 100\%, 50\%, 20\%, and 10\% of training sets. The result indicates that 20\% is the ideal ratio that produces the best performance.

\subsection{Pruning on Images from Minority Races}
\label{sec:minority}

This section examines whether rebalancing the dataset could mitigate pruning-induced bias. Using the UTKFace dataset, where white faces are dominant, we tested SNIP and GraSP with their gradient calculation and parameter selection conducted on non-white examples only (i.e., Black, Asian, Indian) 
Table~\ref{tab:minority} shows the result. 
Interestingly, using a subset of data did not significantly change overall accuracy. However, the overall biases increased compared to the case of using all data. This change shows that the problem of biases in pruned methods cannot be solved by simple data rebalancing and our method effectively addresses this challenging problem.

\begin{table}[h]
\vspace{-8pt}
    \centering
    \resizebox{0.65\linewidth}{!}{
    \begin{tabular}{l| c  c c c c | c  c}
    \multirow{2}{*}{Methods}& \multicolumn{5}{c}{Accuracy} &  \multicolumn{2}{c}{Bias} \\
    \cline{2-8}
     & All  & White & Black & Asian & Indian & $\rho(A)$ & $\rho (\Delta)$ \\
    \hline \hline
     No-pruning & 93.84 & 95.08 & 95.27 & 89.85 & 92.70 & 2.54 & - \\ 
     \hline
  FairGRAPE & 91.72 & 92.86 & 94.18 & 86.88 & 90.40 & 3.21 & 0.78 \\
   GraSP (Minority) & 89.15 & 88.73 & 92.24 & 83.71 & 91.19 & \textcolor{red}{3.80} & \textcolor{red}{2.31} \\  
   GraSP (All data) & 88.33 & 88.80 & 91.05 & 82.47 & 89.13 & 3.73 & 1.77 \\
   SNIP (Minority) & 90.55 & 91.60 & 92.79 & 83.91 & 91.19 & \textcolor{red}{4.03} & \textcolor{red}{1.90} \\ 
   SNIP (All data) & 90.95 & 91.33 & 94.18 & 85.44 & 91.11 & 3.66 & 1.62 \\
    \hline
    \end{tabular}
    }
    \caption{UTKFace gender classification accuracy on minority subsets. $\rho (A)$ and $\rho (\Delta)$ are the standard deviation of accuracy and accuracy loss across racs, respectively.}
    \vspace{-5pt}
    \label{tab:minority}
\end{table}

\subsection{Analysis on Model Sparsity Levels} \label{sec:analysis2}

\begin{figure*}[!t]
    \includegraphics[width=\linewidth,height=0.36\linewidth]{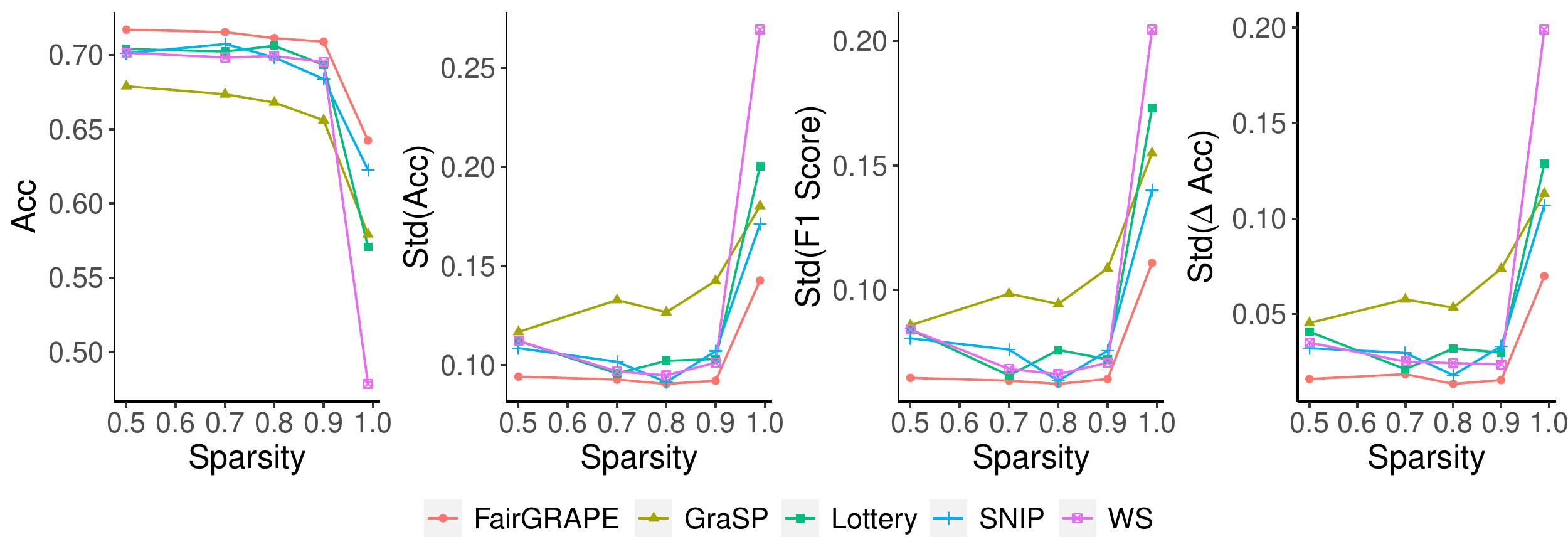} 
     \caption{Accuracy and biases of race classification across races at different sparsity levels. Experiments were conducted using ResNet-34 on FairFace dataset. } 
    \label{fig:acc_vs_sparsity}
\end{figure*}

We next evaluate the performance of FairGRAPE across different sparsity levels to understand its effectiveness.
Figure~\ref{fig:acc_vs_sparsity} shows changes in accuracy and biases over different sparsity levels. 
FairGRAPE outperforms the baseline methods by producing the highest accuracy and lowest disparity of performance degradation across sensitive groups at various pruning rates. As sparsity changes from 90\% to 99\%, most baseline methods exhibit a sharp decrease in accuracy and increase in bias, while performance change in FairGRAPE is substantially smaller. This confirms that our method can be widely deployed to real-world systems with various sparsity levels. 

\subsection{Layer-wise Importance Scores and Bias} \label{sec:analysis3}

\begin{figure}[!t]
  \centering
  \includegraphics[width=1\columnwidth,height=0.4\columnwidth]{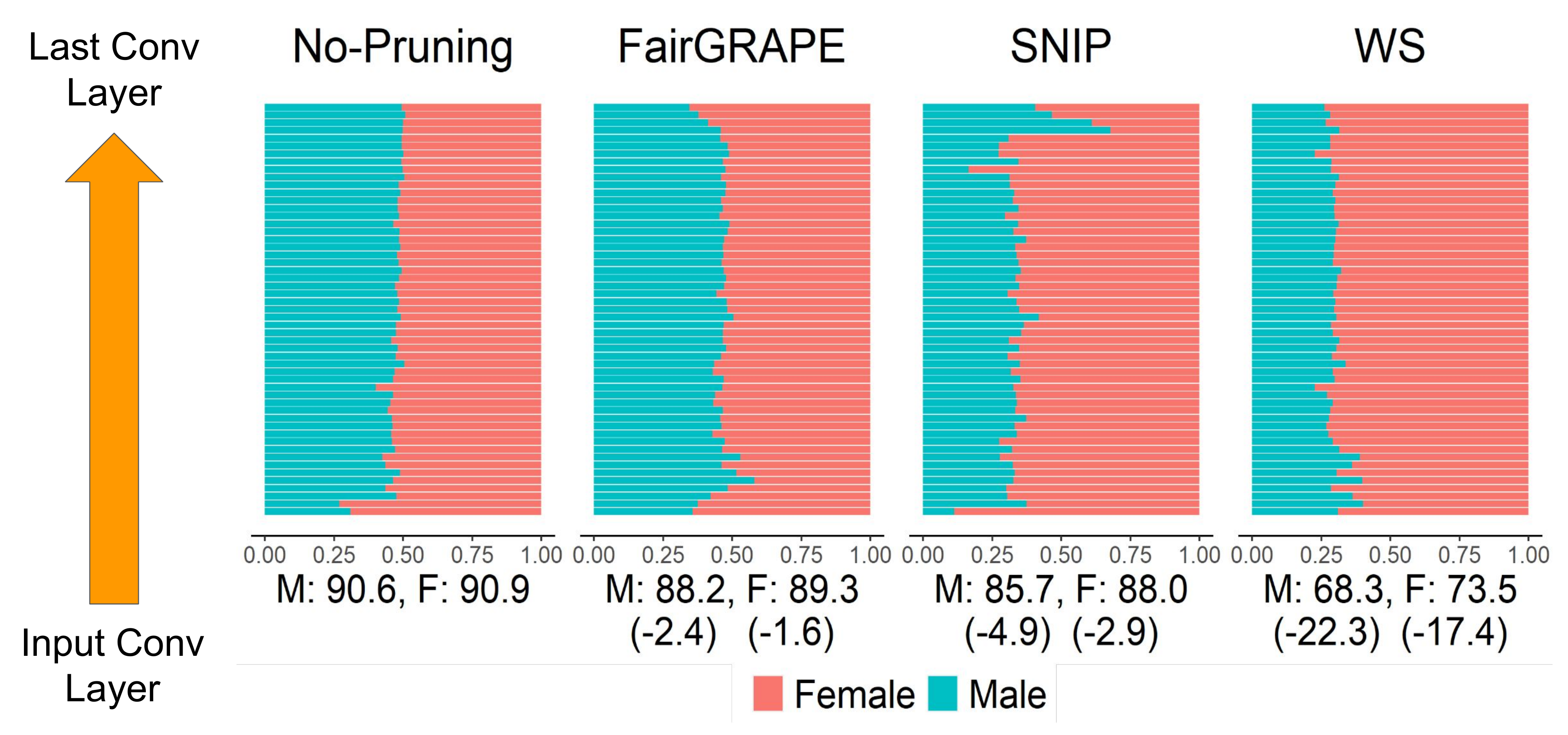} 
  \caption{The ratio of importance scores on MobileNet-V2. Networks are pruned to 90\% sparsity and trained on UTKFace dataset. M and F indicate race classification accuracy on male and female images. Accuracy changes between the pruned models and the full model are shown in parenthesis.} 
  \label{fig:layer}
\end{figure}

This subsection performs an in-depth structural analysis on pruned networks. 
Figure~\ref{fig:layer} visualizes the ratio of importance scores at each layer for each gender group. Each bar represents a convolutional or linear layer. 
and the width of a colored segment indicates the ratio of importance score for the corresponding gender group. 
FairGRAPE preserves the balanced importance distribution of the full network, with similar scores for both genders, leading to substantially smaller gaps in accuracy and accuracy change.
The group-agnostic pruning methods, including SNIP and Weight Selection, select weights with higher importance for the female group, which is already showing higher accuracy in the original model. Consequently, the accuracy of the male group suffered from a substantially greater loss and the gap is much larger than the model pruned by FairGRAPE. 

\section{Conclusion}
In this paper, we proposed FairGRAPE, a novel pruning method that prunes weights based on their importance with respect to each demographic sub-group in the dataset. Empirical results show that our method can minimize performance degradation across sub-groups in different network architectures and datasets at various pruning rates.
We also demonstrated that the association between distributions of gradient importance and performance biases has an important implication for understanding information loss during model compression. 
Our work will therefore contribute to developing fair light-weight models that can be deployed on many mobile devices by mitigating hidden biases. 




\subsubsection{Acknowledgement}
This work was supported by NSF SBE-SMA \#1831848.

%
%
\bibliographystyle{splncs04}
\bibliography{egbib}

\end{document}